\documentclass[journal]{IEEEtran}

\usepackage{amsmath,amssymb,amsfonts}
\usepackage{algorithmic}

\usepackage{graphicx}
	\graphicspath{{figs/}}

\usepackage{textcomp}

    \usepackage{enumitem}
    \usepackage{amssymb}
    \usepackage{amsthm}
    \usepackage{braket}
    \usepackage{supertabular}
    \usepackage{amssymb}
    \usepackage{amsmath,amsthm,amssymb}
    \usepackage{lipsum,graphicx,subcaption}

\usepackage{array}

\begin{document}

\title{Enacted Visual Perception: A Computational Model based on Piaget Equilibrium}

\author{Aref~Hakimzadeh,  
        Yanbo~Xue, 
        and~Peyman~Setoodeh
\thanks{A. Hakimzadeh and P. Setoodeh are with the School
of Electrical and Computer Engineering, Shiraz University, Shiraz,
Iran (e-mail: sa.hakimzadeh@shirazu.ac.ir; psetoodeh@shirazu.ac.ir).}
\thanks{Y. Xue is with the Career Science Lab, Beijing, China, and also with the Department of Control Engineering, Northeastern University 
Qinhuangdao, China (e-mail: yxue@careersciencelab.com).}
}

\maketitle

\begin{abstract}
In Maurice Merleau-Ponty's phenomenology of perception, analysis of perception accounts for an element of intentionality, and in effect therefore, perception and action cannot be viewed as distinct procedures. In the same line of thinking, Alva No\"{e} considers perception as a thoughtful activity that relies on capacities for action and thought. Here, by looking into psychology as a source of inspiration, we propose a computational model for the action involved in visual perception based on the notion of equilibrium as defined by Jean Piaget. In such a model, Piaget's equilibrium reflects the mind's status, which is used to control the observation process. The proposed model is built around a modified version of convolutional neural networks (CNNs) with enhanced filter performance, where characteristics of filters are adaptively adjusted via a high-level control signal that accounts for the thoughtful activity in perception. While the CNN plays the role of the visual system, the control signal is assumed to be a product of mind.
\end{abstract}


\section{Introduction}

\IEEEPARstart{A}{rtificial}  intelligence (AI) can significantly benefit from the ongoing research in neuroscience and psychology. Looking into these fields as a source of inspiration will pave the way for building learning machines that mimic biological brains. CNN inspired by studying cat's visual cortex can be viewed as a success story in following this line of thinking. In the past few years, CNN and its variants have played key roles in building machine-vision systems. CNN-based architectures are built from a combination of layers that implement convolution, pooling, and nonlinear functions \cite{goodfellow2016deep}. In convolutional layers of a CNN, usually, a set of fixed filters are used to scan the input data, hence, there is a lack of controllability over the filters \cite{wu2017introduction}. Adaptively changing the filters at each layer (possibly independent from other layers) will enhance the network's degree of plasticity even without changing the layer structure of the network. Deploying such a mechanism will pave the way for building effective attentive sensors. Let us assume that we are watching a landscape (i.e., a soccer game), sometimes we prefer to focus on a particular region or object (i.e., ball in a soccer game), and sometimes we are interested in a wider view. The basic CNN variants do not provide such a degree of freedom unless they are equipped with an attention mechanism \cite{vaswani2017attention,bello2019attention} or a high-level controller to adjust the shape of filters. If a CNN-based machine-vision algorithm is used in a more sophisticated system such as a robot or an autonomous-driving car, the high-level control signal for adjusting the CNN's filters can be provided by the entity that plays the brain's role and is superior to the visual system.

Human's ability to learn from visual data has unique characteristics that distinguish it from the state-of-the-art machine vision systems. Zoom-in and zoom-out are common procedures that we do every day. Most of the time we do not scan our environment precisely until something raises our curiosity. Biological visual systems use a fixed number of photoreceptors in the retina to convert light into nerve impulses that are transmitted to the higher layers of the visual cortex. While the main neural structure involved in vision (in the sense of the number of photoreceptors) remains unchanged for different manners of looking, visual perception occurs through moving eyes around (saccade) and zooming. Hence, visual perception can be better understood as an enacted perception \cite{merleau1996phenomenology,noe2004action}. Here, we aim at designing a system that is able to mimic human adaptive visual behavior despite structural constraints. A control signal is used to control how a scene should be screened or what part of the scene should be focused on. For instance, a control variable is needed to determine the manner of focusing on a certain object or a specific area to detect movements or anomalies. To address this issue, we look into psychology as a source of inspiration to define a proper mind status that can be interpreted as the required control signal \cite{raab2005cognition}. In what follows, more details will be provided on implementing the control mechanism for adjusting filters followed by proposing a criterion for equilibrium. Then, the proposed model will be validated on a real dataset.

The rest of the paper is organized as follows. Section~II reviews the Piaget's schema theory and the equilibrium concept. Section~III covers the filter design for CNN. Section~IV provides simulation results. Section~V presents the concluding remarks.

\section{Piaget's Equilibrium}

As mentioned previously, in order to have a precise definition of the control signal for guiding the visual system, it would be helpful to use Piaget's equilibrium. First, we recall the definitions of schema, assimilation, accommodation, and equilibration from psychology \cite{piaget2008psychology}:
\begin{itemize}
\item 	\textit{Schemas} refer to the basic building blocks of cognition that make it possible to form a mental representation of the environment. 
\item 	\textit{Assimilation} refers to the similarity between the existing schemas and a new situation encountered. 
\item 	\textit{Accommodation} refers to the elements of a new situation that are either not contained in the existing schema or contradict them.
\item 	\textit{Equilibration} refers to the balance between assimilation and accommodation. Having the intention to perform a task or find a solution to a problem calls for assimilation of information to partially match the individual's mental schemas as well as accommodation of information by modifying the individual's way of thinking to adopt it. Therefore, problem solving can be studied under an equilibration criterion \cite{singer1997piaget}.  
\end{itemize}

The notion of equilibration can be adopted in CNN-based machine vision systems, where a control signal is needed to reflect mind's status. In this framework, for filters with fixed numbers of cells, the control signal will be responsible for cells' topology arrangement. There is no need for such filters to be fully connected, which means that there could be a gap between filter's cells. To be more precise, some of the filter's cells could be null cells. According to Piaget's definition of equilibrium, as the distance from the equilibrium increases in a non-equilibrium condition, filter cells should be more condensed and filter should sweep the image faster till the distance decreases. Then, in a close-to-equilibrium condition, a sparser filter that sweeps the image slower would work. The sweeping speed of the filter can be adjusted by its stride. The measure of distance from equilibrium can be considered as a value between zero and one, where zero reflects a completely stable mind and one means a totally unstable mind. Such mechanisms can be embedded into the CNN architecture using floating discrete controllable filters as described in the following section.

\section{Filter Realization}

\begin{figure}
\begin{center}
\includegraphics[width=1\linewidth]{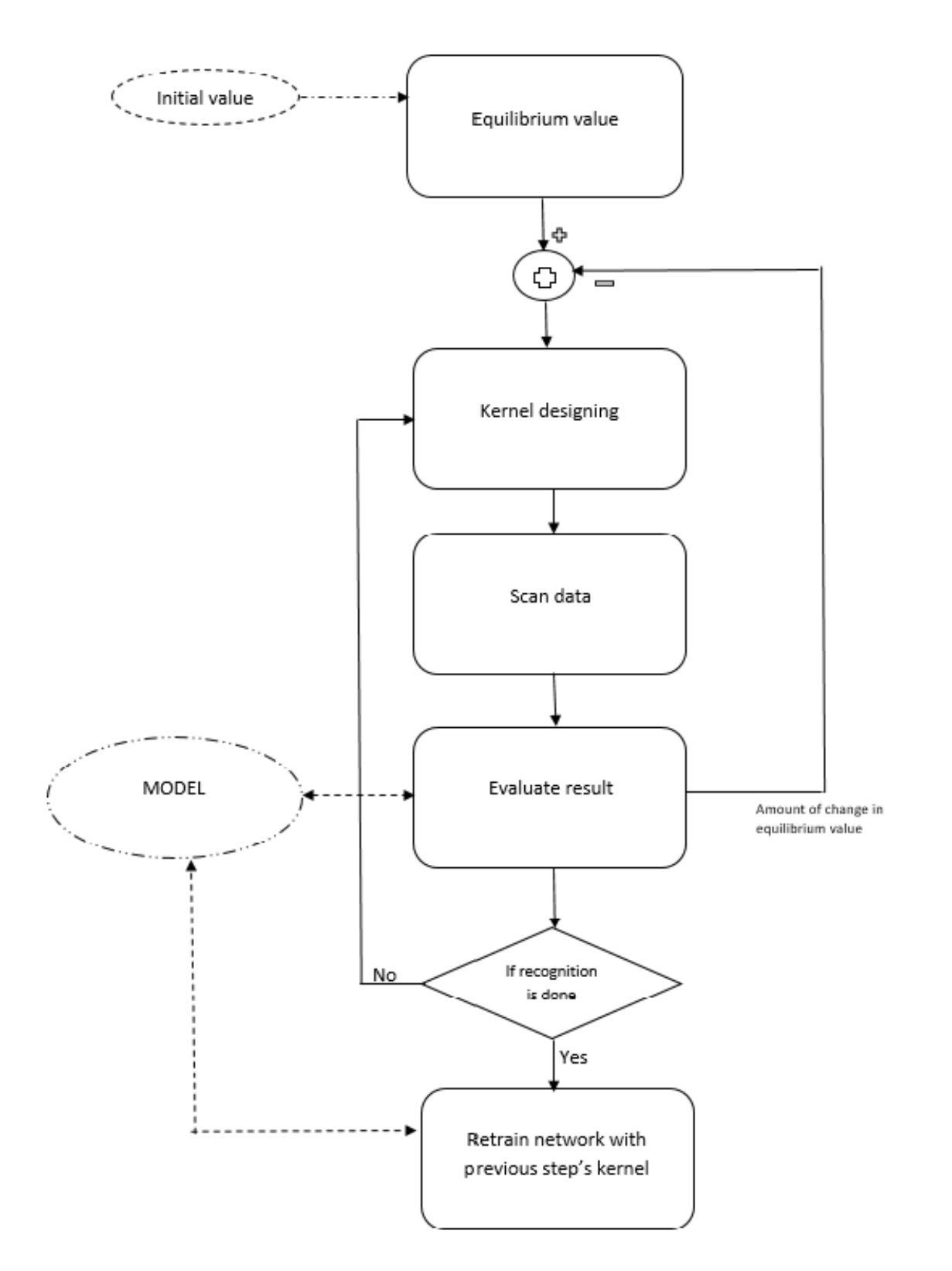}    
\caption{Training procedure for a CNN with floating discrete controllable filters.} 
\label{CNNDCF}
\end{center}
\end{figure}

Flowchart of the training procedure for a CNN with floating discrete filters (FDF) is depicted in Fig. \ref{CNNDCF}. Two approaches are proposed for filter implementation using neural networks and mathematical functions:
\begin{itemize}
\item Mathematical function: A family of spiral-shaped functions with adjustable interim and boundaries can be used for filter realization. In polar coordinates, such filters are mathematically described as $f(\theta)=R(\theta) e^{j\theta}$, where radius $R$ is a function of phase $\theta \in [0 , 2k\pi]$, and k is chosen according to the step size used for discretization as well as the total number of filter's cells. Choosing the form of the function $R(\theta)$ provides a degree of freedom in the design process. One option would be $R(\theta)=\frac{\theta^{\alpha n}}{\beta}$, where $n \in (-1 ,  1)$ denotes the scaled equilibrium value, $\alpha \in (0 ,  1)$ is a hyper-parameter, and $\beta$ is a normalizing coefficient. While for negative values of $n$, filter cells are more concentrated around the center, for positive values of $n$, they are more scattered. Hence, adjusting the value of $n$ may be viewed as taking a countermeasure to compensate for the distance from equilibrium. Fig. 2 shows a number of filters plotted for different values of $n$ with $\alpha= \beta=1$. It should be noted that in general, $\alpha$ and $\beta$ will have different values.
\item Neural network: A multilayer perceptron (MLP) \cite{haykin2010neural,gurney1997introduction} can be trained for filter realization. In each computation step, the MLP receives the equilibrium value as its input and returns the location of 25 percent of the cells in the filter with their corresponding values as its output. The other 75 percent of the cells are considered as null cells.  
\end{itemize}

\begin{figure}
\begin{center}
\includegraphics[width=1\linewidth]{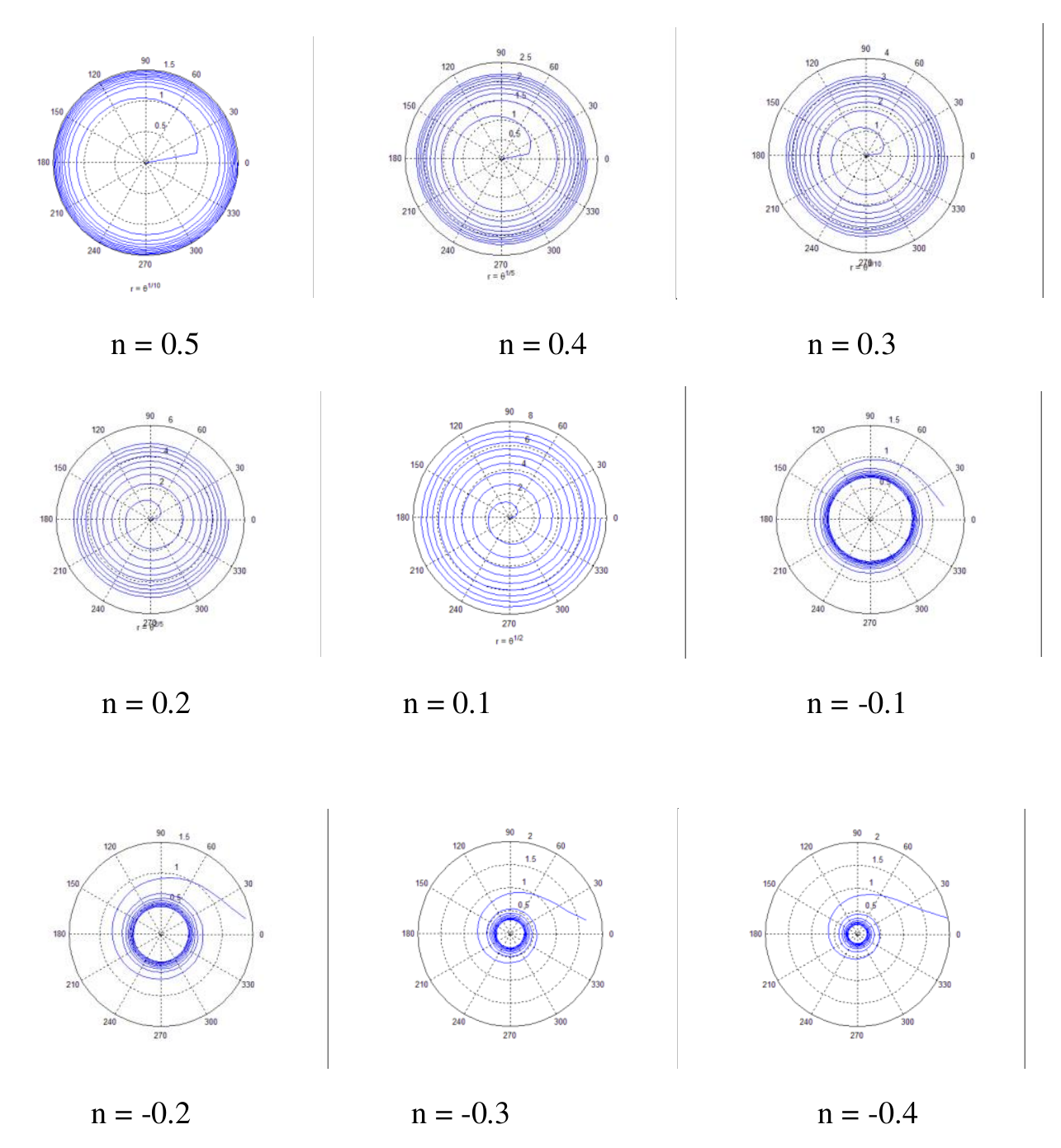}    
\caption{Filter shapes $f(\theta)=R(\theta) e^{j\theta}$  for different values of n with $\alpha= \beta=1$. As the distance from equilibrium increases, we will have a more concentrated filter around center.} 
\label{Filters}
\end{center}
\end{figure}

One common technique in machine learning is feature encoding. Therefore, it would be informative to investigate the proposed algorithm from the perspective of hierarchical data abstraction. Although the number of cells in a filter may be constant during visual data processing with floating discrete filters, at each step, the information stored in weights have different interpretations. Every equilibrium value leads to a different filter shape, and therefore, a different representation of the image in the form of filter weights. As the distance from equilibrium decreases and filters with more scattered shapes can be used, weights of the network will provide a more abstract representation of the image. In different applications, every set of weights, which corresponds to an equilibrium value, can be separately stored in memory and used for decoding data at various levels of abstraction. A relational learning scheme can be deployed for learning a bidirectional association between different equilibria and filter shapes. Using such a memory-augmented version of the algorithm \cite{graves2016hybrid}, discrete floating filters can be used for encoding visual data, and equilibrium values provide the key for decoding data.

To expedite the learning process, we can start with a pre-trained network using the most scattered filter possible. If results are not satisfying, distance from equilibrium increases and filter's shape is changed accordingly. This characteristic allows for using the proposed architecture for anomaly detection. Moreover, it paves the way for deriving attentive algorithms. In a recognition system, in the last layer before the softmax layer in the CNN, we can determine the best part of the data to pay attention to. The problem of augmenting the algorithm with an attention mechanism can be divided into two major sub-problems:
\begin{itemize}
\item Class-based attention: According to \cite{zhou2016learning}, attention can be implemented using the pooling operation in the last layer of the CNN. The class activation mapping uses the weights connected to the nodes associated to the determined classes in the output layer. Hence, in the last layer of the CNN, dimension can be matched to the input dimension using an up-sampling mechanism. The same approach can be adopted by the proposed algorithm followed by re-scanning the highlighted part with a more condensed filter. This process resembles how humans find a region of interest in a scene and then focus on it.
\item Attention-based anomaly detection: The most relevant algorithm to this idea is the grad-CAM \cite{selvaraju2017grad}. In a pre-trained network, a change that occurs in the input increases the gradient in the last layer with respect to the previous knowledge stored in the memory. Back-propagating these gradients would cause some nodes of the last layer of the CNN to become more influential than others. By up-sampling the output to match the size of the input, the area that caused an increase in the distance from equilibrium or the part containing abnormality can be determined. Then, a decision can be reached using more condensed filters on the specified area. It is similar to the class-based attention with the difference that here, the gradient of the softmax layer is tracked by the system instead of the weights connected to the specific output node.
\end{itemize}

\section{Experiments}

Floating discrete filters can be designed by solving the following bi-level optimization problem:
\begin{eqnarray}
\label{eq:bilevel} {\underset {}{\text{maximize}}} & & \;
O
\\
\text{subject to:}  &\:& {\underset {W}{\text{minimize}}}  \;
\sum_i (y_i^d -y_i)^2  \nonumber
 \\
&\:& \text{subject to:}   \; \; \; \; equilibrium > c \nonumber
\end{eqnarray}
where $O$ denotes the order of the current filter, $y^d$ and $y$ are the desired output and the network output scanned by a filter of order $O$, respectively. $c$ is a constant that determines the threshold for equilibrium value (i.e., a measure of distance from equilibrium condition), and $W$ refers to network parameters. This problem can be solved in an iterative manner; starting with a value for the manipulated variable of the leader problem, the follower problem is solved, then, fixing the manipulated variable of the follower problem, the leader problem is solved to find a better value for the manipulated variable of the leader problem. In this way, both of the manipulated variables converge to their optimal values. To design an optimal filter, we start with the most scattered structure, which corresponds to the highest possible order of the filter. If the proposed order satisfies the follower problem, process ends; otherwise, order is decreased and the follower problem is checked until both levels of the problem are satisfied.

The CIFAR-10 benchmark dataset was chosen to evaluate the proposed architecture, where 60\%, 20\%, and 20\% of data points were selected as the training, validation, and test sets, respectively. In order to implement the idea of floating discrete filters, the notion of Piaget's equilibrium must be defined mathematically. For instance, equilibrium can be defined as a function of variance of the network output, such that a higher variance represents a larger distance from equilibrium. Three sets of experiments were performed. In the first experiment, fulfilling the required conditions, the deployed floating discrete filters were handcraft designed for each discrete equilibrium value. In this scenario, the kernel design step in the flowchart of Fig. \ref{CNNDCF} can be bypassed. In the second experiment, filters were designed using the mathematical functions described in the previous section for given equilibrium values.

These experiments were performed as follows. In the first experiment, a network was trained with normal filters, and the trained network's performance was evaluated based on the test set. In the second experiment, equilibrium was initially set to the highest possible value (one). When network makes a prediction using designed filters, if variance of the output layer is acceptable, process terminates, otherwise, equilibrium value is decreased. Decreasing the equilibrium value results in more concentrated filters.  Then, network output is calculated based on the new filters, and this procedure continues until the network can provide a trustworthy answer, which can be interpreted as reaching the desired threshold for variance. 
Due to discretization, there exist only nine possible equilibrium configurations and filter shapes, hence, there will be nine possible steps in the FDF design process.

The following four criteria are considered to compare the performance of different architectures: 
\begin{itemize}
\item Criterion 1: percentage of the correct answers predicted by FDFs (algorithm precision)
\item Criterion 2: normal filter prediction, which is availabke as a candidate during the FDF design process
\item Criterion 3: first and last predicted answers of a single FDF process are the same
\item Criterion 4: mean value of the number of times that filter shape was changed during the process (mean number steps per instance)
\end{itemize}

In the first experiment, three different networks with different architectures were trained. For each network, six different thresholds were considered for variance, which determines the stopping criterion, and for each threshold, the mentioned four criteria were calculated as percentage. The first architecture has the following characteristics:
\begin{itemize}
\item First layer: a CNN with 3x3 filter size and one output filter
\item Second layer: a fully connected layer with 1024 neurons
\item Output layer: a layer with 10 neurons
\end{itemize}
Results are summarized in Table \ref{tbl:table1}. Precision of this network with normal filter shape was 43.99\%. Number 9 appearing in the last row for criterion 4 shows that uncertainty was present till the designed filter took a shape similar to the normal filter. This condition can be interpreted as reaching zero value for equilibrium. First column of the table shows a precision close to that of normal filter while the mean number of steps for each instance is near to one. This means that using a filter, which is twice larger than the normal one and reduces the computational burden in each layer up to 50\%, could achieve a precision close to the typical process. Considering only the first output of the FDF process and ignoring the rest, which is equivalent to designing the filter with an equilibrium value of one, leads to similar predictions for both FDF and normal filter 83.81\% of the time.  This confirms the generality of designed filters regadless of the process functionality, which is reflected in criterion 3.

The second architecture used in the first experiment has the following characteristics:
\begin{itemize}
\item First layer: a CNN with 3x3 filter size with three output filter
\item Second layer: a fully connected layer with 1024 neurons
\item Output layer: a layer with 10 neurons
\end{itemize}
Results are summarized in Table \ref{tbl:table2}. Precision of this network with normal filter shape was 52.47\%. In this table, the second column has benefits of having near normal filter precision and small mean number of steps, which confirms reduced computational effort while achieving nearly the same level of performance. Regarding generalization, by just considering the first-stage prediction of FDF, in 72.47\% of the time, predictions would be the same as those of the normal filter.

The third architecture used for the first experiment has the following characteristics:
\begin{itemize}
\item First layer: a CNN with 3x3 filter size with eight output filter
\item Second layer: a fully connected layer with 1024 neurons
\item Output layer: a layer with 10 neurons
\end{itemize}
Results are summarized in Table \ref{tbl:table3}. Precision of this network with normal filter shape was 57.9\%. Similar to the previous two cases, for this network, reduced computational burden leads to faster performance.

For the second experiment, an architecture similar to the first architecture was used and a kernel was designed based on the mathematical functions described in the previous section. The following architecture was used for this experiment:
\begin{itemize}
\item First layer: a CNN with 3x3 filter size with one output filter
\item Second layer: a fully connected layer with 1024 neurons
\item Output layer: a layer with 10 neurons
\end{itemize}
Results are summarized in Table \ref{tbl:table4}. By comparing tables \ref{tbl:table1} and \ref{tbl:table4}, which correspond to the same architecture, we can conclude that the mathematical filter design approach shows promising results and achieves normal filter precision with less computational effort. The third row of Table \ref{tbl:table4} shows the generality of this approach, which can be used for data compression or even abstraction without loss of information. 

Results show that increasing the complexity of the network will increase the gap in performance between architectures that use FDFs and normal filters. It was expected because the FDF process was designed to improve the generalization ability of CNNs via keeping a simple structure and providing flexibility to filters, while maintaining an acceptable level of performance. However, increasing complexity is against the main idea behind the FDF design process. Improving performance of FDF must be achieved through better definition of the equilibrium value. In this section, a simple interpretation of equilibrium in terms of variance was implemented to show the effectiveness of the proposed method.

\begin{table*}[h!]
\begin{center}
    \caption{Results of experiment 1, case 1.}
    \label{tbl:table1}
  \begin{tabular}{c | c c c c c c}
\hline
    Variance Threshold & 0.05 & 0.07 & 0.08 & 0.088 & 0.09 & 0.1 \\ 
 \hline
   Criterion 1 & 42.06 & 41.53 & 41.23 & 40.69 & 43.99 & 43.99 \\
   Criterion 2 & 84.60 & 85.31 & 85.86 & 87.06 & 99.56 & 99.56 \\ 
   Criterion 3 & 95.54 & 91.43 & 88.74 & 83.68 & 83.93 & 83.93 \\ 
   Criterion 4 & 1.089 & 1.227 & 1.356 & 1.703 & 9 & 9 \\ 
 \hline
   \end{tabular}
\end{center}
\end{table*}

\begin{table*}[h!]
\begin{center}
    \caption{Results of experiment 1, case 2.}
    \label{tbl:table2}
  \begin{tabular}{c | c c c c c c}
 \hline
    Variance Threshold & 0.05 & 0.07 & 0.08 & 0.088 & 0.09 & 0.1 \\ 
 \hline
   Criterion 1 & 34.38	& 48.95 &	48.57 & 48.98 & 49.42 & 52.74 \\
   Criterion 2 & 41.12	& 76.46 & 78.02 & 81.96 & 86.86 & 99.49 \\ 
   Criterion 3 & 90.03	& 86.22 & 83.1	& 79.06 & 76.09 & 72.37 \\ 
   Criterion 4 & 1.36 & 1.756 & 2.42 & 3.412 & 4.714 & 9 \\ 
 \hline
   \end{tabular}
\end{center}
\end{table*}

\begin{table*}[h!]
\begin{center}
    \caption{Results of experiment 1, case 3.}
    \label{tbl:table3}
  \begin{tabular}{c | c c c c c c}
\hline
    Variance Threshold & 0.05 & 0.07 & 0.08 & 0.088 & 0.09 & 0.1 \\ 
\hline
   Criterion 1 & 51.23 & 50.34 & 49.7 & 49.02 & 	57.9 & 57.9 \\
   Criterion 2 & 75.0 & 76.19 & 76.99 & 79.19 & 	99.44 & 99.44 \\ 
   Criterion 3 & 94.13 & 89.56 & 86.54 & 81.28 & 73.51 & 73.51 \\ 
   Criterion 4 & 1.185 &	1.448 & 1.719 & 2.465 & 9 & 9 \\ 
\hline
   \end{tabular}
\end{center}
\end{table*}

\begin{table*}[h!]
\begin{center}
    \caption{Results of experiment 2.}
    \label{tbl:table4}
  \begin{tabular}{c | c c c c c c}
\hline
    Variance Threshold & 0.05 & 0.07 & 0.08 & 0.088 & 0.09 & 0.1 \\ 
\hline
   Criterion 1 & 43.71	& 43.61 & 43.62 & 43.45 & 43.37 & 43.99 \\
   Criterion 2 & 82.25 & 83.26 & 84.03 & 85.88 & 89.39 & 99.32 \\ 
   Criterion 3 & 95.7 & 92.42 & 90.59 & 87.02 & 	83.43 & 81.16 \\ 
   Criterion 4 & 1.126 & 1.333 & 1.543 & 2.12 & 3.32 & 9 \\ 
\hline
   \end{tabular}
\end{center}
\end{table*}

\begin{table*}[h!]
\begin{center}
    \caption{Results of experiment 3, case 1.}
    \label{tbl:table5}
  \begin{tabular}{c | c c c c c c}
\hline
    Entropy Threshold & 10e-11 & 0.001 & 0.005 & 0.02 & 0.2 & 0.3 \\ 
\hline
   Criterion 1 & 45.0 & 44.3 & 43.7 & 42.9 & 40.9 & 41.8 \\
   Criterion 2 & 80.8 & 81.1 & 81.4 & 82.7 & 84.7 & 89.5 \\ 
   Criterion 3 & 100 & 97.9 & 96.1 & 90.2 & 82.0 & 80.4 \\ 
   Criterion 4 & 1.03 & 2.02 & 2.41 & 3.04 & 3.88 & 5.021 \\ 
\hline
   \end{tabular}
\end{center}
\end{table*}

\begin{table*}[h!]
\begin{center}
    \caption{Results of experiment 3, case 2.}
    \label{tbl:table6}
  \begin{tabular}{c | c c c c c c}
\hline
    Entropy Threshold & 10e-11 & 0.001 & 0.005 & 0.02 & 0.2 & 0.3 \\ 
\hline
   Criterion 1 & 42.6 & 41.12 & 39.3 & 38.3 & 38.5 & 40.0 \\
   Criterion 2 & 84.45 & 84.45 & 83.5 & 83.9 & 85.6 & 90.0 \\ 
   Criterion 3 & 99.75 & 95.87 & 90.5 & 83.1 & 77.7 & 77.8 \\ 
   Criterion 4 & 1.053 & 1.47 & 2.04 & 2.59 & 3.45 & 4.38 \\ 
\hline
   \end{tabular}
\end{center}
\end{table*}

In the third experiment, entropy of the output layer was used instead of variance to represent equilibrium, such that a higher entropy demonstrates a less confident prediction and hence a lower equilibrium value \cite{cognitivecontrol2012, fatemi2017observability}. Two case studies were investigated with network architectures and configurations similar to the first case in the first experiment and the second experiment except for using entropy instead of variance. Results are summarized in tables \ref{tbl:table5} and \ref{tbl:table6}, respectively. Networks can achieve better precision and faster processing in this case. In addition to improvement in the processing time, using entropy to represnt equilibrium improved precision even beyond what was achievable by the CNN with normal filters using variance (Tables \ref{tbl:table5} and \ref{tbl:table6} cover a wider range of equilibrium thresholds compared to tables \ref{tbl:table1} to \ref{tbl:table4}). Therefore, entropy can provide a valid stopping criterion during thr training phase. 

Although the two proposed mathematical representations of equilibrium were fairly simple, experiments showed the effectiveness of the proposed approach for improving the flexibility of CNNs. Hence, it can be viewed as a step towards designing a proper high-level controller for mimicking human visual system. 

\section{Concluding Remarks}

Performance of a machine vision system can be enhanced by rescanning data using a subset of filters after the answer is known in order to fine-tune some of the filter weights. In this way, the system can recognize the picture with fewer iterations and a wider filter. Moreover, using the outcome of the dense layers to adjust the metric for the distance from equilibrium, may improve the performance. As an attempt to provide a computational model for Piaget's original definition of equilibrium, two mathematical measures were deployed in the implementations and their results were compared. Implementing different controlling mechanisms in a machine vision system with any existing network architecture is now possible through different definitions of equilibrium value. This paves the way for using any CNN variant as a follower optimizer for another higehr-level system, such as an artificial mind. This paper proposed a computational model for the enacted visual perception based on CNNs that deploy floating discrete controllable filters. The design method proposed a cognitive computing architecture that integrates perception, memory, and attention. Deploying a high-level control signal that reflects the mind's status allows the proposed architecture to closely mimic biological vision systems. Moreover, the proposed visual data processing system can be easily integrated in more sophisticated systems, where an entity plays the role of brain. However, training this system may take a longer time to converge and accuracy may be slightly worse than the CNN equivalents with fixed filters. Considering both advantages and disadvantages, the proposed architecture would provide a potential candidate for the machine vision systems deployed in robotic projects aimed at mimicking humans.  

\bibliographystyle{IEEEtran}
\bibliography{IEEEabrv,Main}

\end{document}